\documentclass{article}
\pdfoutput=1
\usepackage[a4paper, total={6in, 9in}]{geometry}

\usepackage[utf8]{inputenc}
\usepackage[T1]{fontenc}
\usepackage{lmodern}
\usepackage{hyperref}
\usepackage{url}
\usepackage{booktabs}
\usepackage{amsfonts}
\usepackage{nicefrac}
\usepackage{microtype}
\usepackage{color}
\usepackage[]{graphicx}
\usepackage{amsmath}
\usepackage{xcolor}
\usepackage{paralist}
\usepackage{enumerate}
\usepackage[]{graphicx}
\usepackage[]{subcaption}

\usepackage{titling}
\usepackage{authblk}

\usepackage{threeparttable}
\usepackage{tikz}
\usepackage{multirow}

\newcommand*\circled[1]{\tikz[baseline=(char.base)]{\node[shape=circle,draw,inner sep=2pt] (char) {#1};}}
            
\usepackage{array}
\newcolumntype{C}[1]{>{\centering\let\newline\\\arraybackslash\hspace{0pt}}m{#1}}
\renewcommand{\arraystretch}{1.2}
\usepackage{caption} 
\newcommand{\draft}[1]{}

\pretitle{\begin{center} \LARGE \bf}
\posttitle{\par\end{center}\vskip 0.5em}
\preauthor{
    \begin{center}
    \normalsize \lineskip 0.5em%
    \begin{tabular}[t]{c}
}
\postauthor{\end{tabular}\par\end{center}}

\predate{\begin{center}\large}
\postdate{\par\end{center}}

\title{Relational Deep Reinforcement Learning}

\author{Vinicius Zambaldi\thanks{Equal contribution.}}
\author{David Raposo\protect\footnotemark[1]}
\author{Adam Santoro\protect\footnotemark[1]}
\author{Victor Bapst}
\author{Yujia Li}
\author{Igor Babuschkin}
\author{Karl Tuyls}
\author{David Reichert}
\author{Timothy Lillicrap}
\author{Edward Lockhart}
\author{Murray Shanahan}
\author{Victoria Langston}
\author{Razvan Pascanu}
\author{Matthew Botvinick}
\author{Oriol Vinyals}
\author{Peter Battaglia}

\affil{
    Contact: \texttt{\normalsize vzambaldi@google.com, draposo@google.com, adamsantoro@google.com}
}
\affil{DeepMind\\London, United Kingdom}

\date{}

\begin{document}
\maketitle

\begin{abstract}

We introduce an approach for deep reinforcement learning (RL) that improves upon the efficiency, generalization capacity, and interpretability of conventional approaches through structured perception and relational reasoning. It uses self-attention to iteratively reason about the relations between entities in a scene and to guide a model-free policy. Our results show that in a novel navigation and planning task called Box-World, our agent finds interpretable solutions that improve upon baselines in terms of sample complexity, ability to generalize to more complex scenes than experienced during training, and overall performance. In the StarCraft II Learning Environment, our agent achieves state-of-the-art performance on six mini-games -- surpassing human grandmaster performance on four. By considering architectural inductive biases, our work opens new directions for overcoming important, but stubborn, challenges in deep RL.

\end{abstract}

\begingroup
\let\clearpage\relax
\section{Introduction}

Recent advances in deep reinforcement learning (deep RL) \cite{mnih2015human, silver2016mastering,RusuVRHPH17} are in part driven by a capacity to learn good internal representations to inform an agent's policy. Unfortunately, deep RL models still face important limitations, namely, low sample efficiency and a propensity not to generalize to seemingly minor changes in the task \cite{garnelo2016towards, zhang2018study, lake2017building, kansky2017schema}. These limitations suggest that large capacity deep RL models tend to overfit to the abundant data on which they are trained, and hence fail to learn an abstract, interpretable, and generalizable understanding of the problem they are trying to solve.

Here we improve on deep RL architectures by leveraging insights introduced in the RL literature over $20$ years ago under the Relational RL umbrella (RRL, \cite{DzeroskiRB98,DzeroskiRD01}). RRL advocated the use of \textit{relational} state (and action) space and policy representations, blending the generalization power of relational learning (or inductive logic programming) with reinforcement learning. We propose an approach that exploits these advantages concurrently with the learning power afforded by deep learning. Our approach advocates learned and reusable entity- and relation-centric functions \cite{battaglia2016interaction, raposo2017discovering, anon2018relational} to implicitly reason \cite{santoro2017simple} over relational representations. 

Our contributions are as follows: (1) we create and analyze an RL task called Box-World that explicitly targets relational reasoning, and demonstrate that agents with a capacity to produce relational representations using a \textit{non-local} computation based on attention \cite{vaswani2017attention} exhibit interesting generalization behaviors compared to those that do not, and (2) we apply the agent to a difficult problem -- the StarCraft II mini-games \cite{vinyals2017starcraft} -- and achieve state-of-the-art performance on six mini-games.

\begin{figure}
    \centering
    \includegraphics[width=.9\textwidth]{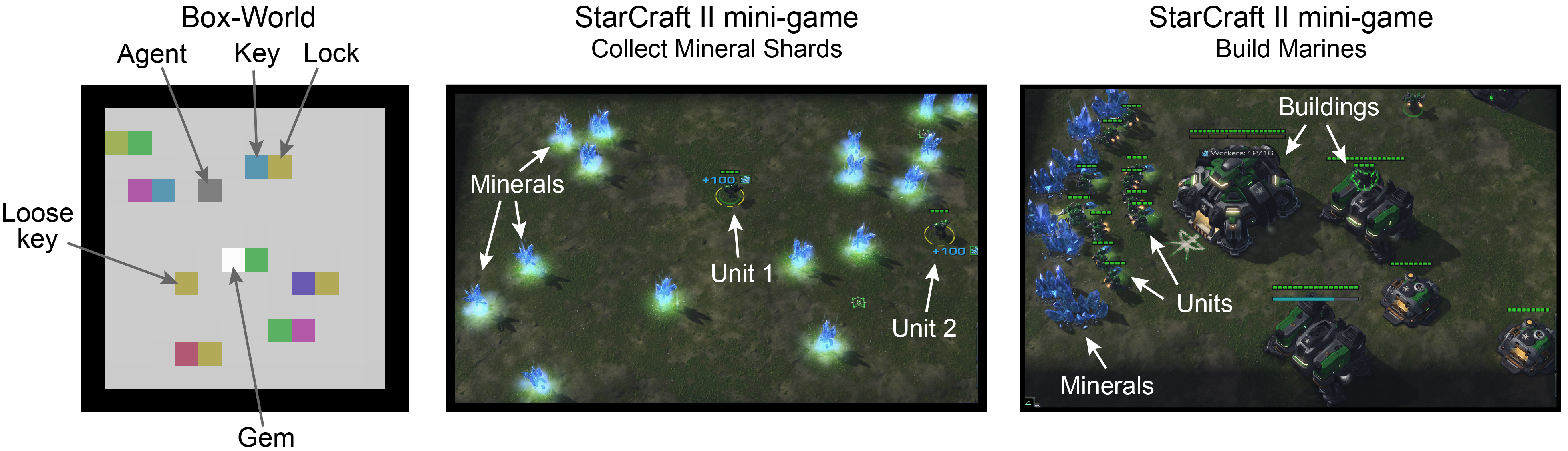}
    \caption{Box-World and StarCraft II tasks demand reasoning about entities and their relations.}
    \label{fig:bw_sc_example}
\end{figure}

\section{Relational reinforcement learning}

The core idea behind RRL is to combine reinforcement learning with relational learning or Inductive Logic Programming \cite{MuggletonR94} by representing states, actions and policies using a first order (or relational) language \cite{DzeroskiRB98,DzeroskiRD01,DriessensR03,DriessensD04}. Moving from a propositional to a relational representation facilitates generalization over goals, states, and actions, exploiting knowledge learnt during an earlier learning phase. Additionally, a relational language also facilitates the use of background knowledge. Background knowledge can be provided by logical facts and rules relevant to the learning problem.

For example in a blocks world, one could use the predicate $\mathit{above}(S, A, B)$ to indicate that block $A$ is above block $B$ in state $S$ when specifying background knowledge. Such predicates can then be used during learning for blocks $C$ and $D$, for example. The representational language, background, and assumptions form the inductive bias, which guides (and restricts) the search for good policies. The language (or declarative) bias determines the way concepts can be represented.

Neural nets have traditionally been associated with the attribute-value, or propositional, RL approaches \cite{Otterlo}. Here we translate ideas from RRL into architecturally specified inductive biases within a deep RL agent, using neural network models that operate on structured representations of a scene -- sets of entities -- and perform relational reasoning via iterated, message-passing-like modes of processing. The entities correspond to local regions of an image, and the agent learns to attend to key objects and compute their pairwise and higher-order interactions.

\section{Architecture}

We equip a deep RL agent with architectural inductive biases that may be better suited for \textit{learning} (and computing) relations, rather than specifying them as background knowledge as in RRL. This approach builds off previous work suggesting that relational computations needn't necessarily be biased by entities' spatial proximity \cite{wang2017non, battaglia2016interaction, watters2017visual, raposo2017discovering, santoro2017simple,hu2017relation}, and may also profit from iterative structured reasoning \cite{bello2016neural,dai2017learning,mishra2017simple,kool2018attention}. 

Our contribution is founded on two guiding principles: \textit{non-local computations} using a shared function and \textit{iterative computation}. We show that an agent which computes pairwise interactions between entities, independent of their spatial proximity, using a shared function, will be better suited for learning important relations than an agent that only computes local interactions, such as in translation invariant convolutions\footnote{Intuitively, a ball can be related to a square by virtue of it being ``left of'', and this relation may hold whether the two objects are separated by a centimetre or a kilometer.}. Moreover, an iterative computation may be better able to capture higher-order interactions between entities.

\begin{figure}
    \centering
    \includegraphics[width=1\textwidth]{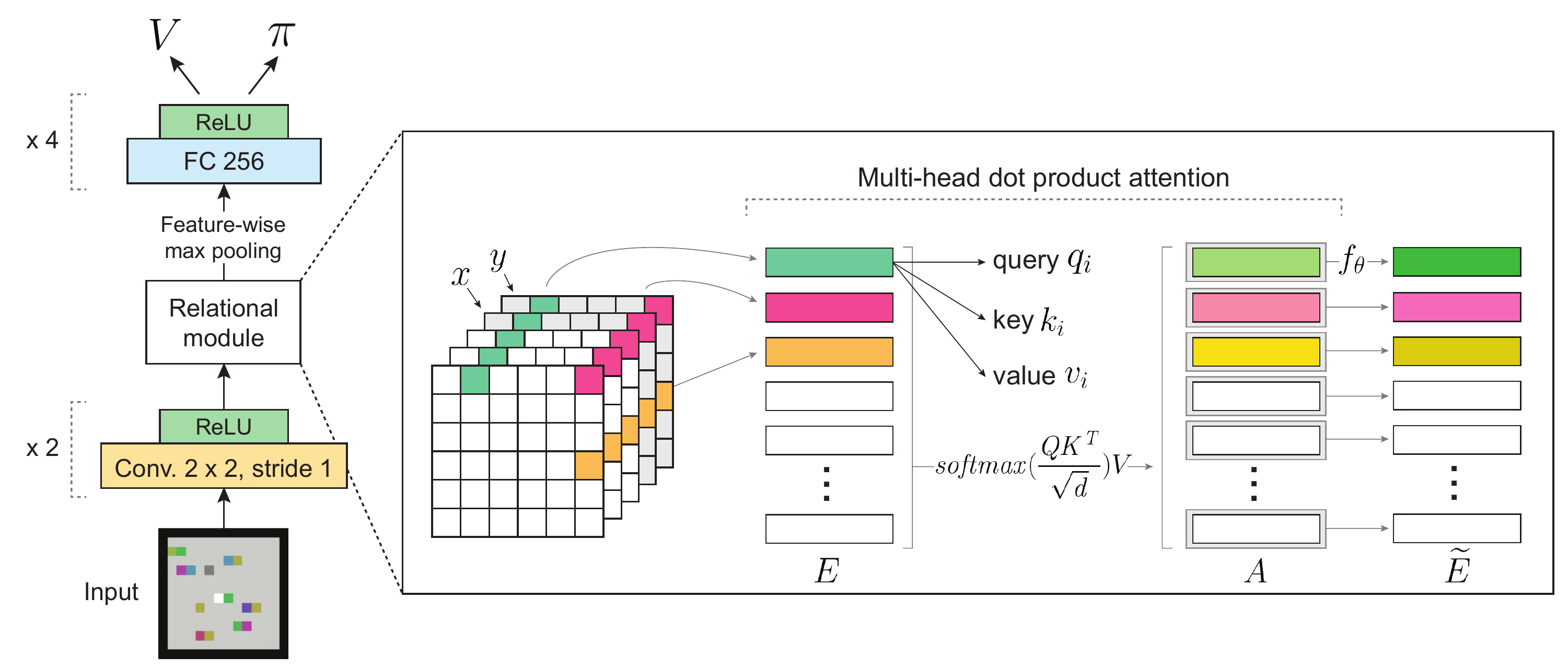}
    \caption{Box-World agent architecture and multi-head dot-product attention. $E$ is a matrix that compiles the entities produced by the visual front-end; $f_{\theta}$ is a multilayer perceptron applied in parallel to each row of the output of an MHDPA step, $A$, and producing updated entities, $\widetilde{E}$.}
    \label{fig:agent_bw}
\end{figure}

\subsubsection*{Computing non-local interactions using a shared function}

Among a family of related approaches for computing non-local interactions \cite{wang2017non}, we chose a computationally efficient attention mechanism. This mechanism has parallels with graph neural networks and, more generally, message passing computations \cite{scarselli2009graph, niepert2016learning, kipf2016semi,anon2018relational,denil2017programmable}. In these models entity-entity relations are explicitly computed when considering the messages passed between connected nodes of the graph.

We start by assuming that we already have a set of entities for which interactions must be computed. We consider multi-head dot-product attention (MHDPA), or \textit{self-attention} \cite{vaswani2017attention}, as the operation that computes interactions between these entities.

For $N$ entities ($\mathbf{e}_{1:N}$), MHDPA projects each entity $i$'s state vector, $\mathbf{e}_i$, into query, key, and value vector representations: $\mathbf{q}_i$, $\mathbf{k}_i$, $\mathbf{v}_i$, respectively, whose activities are subsequently normalized to have $0$ mean and unit variance using the method from \cite{ba2016layer}. Each $\mathbf{q}_i$ is compared to all entities' keys $\mathbf{k}_{1:N}$ via a dot-product, to compute unnormalized saliencies, $\mathbf{s}_i$. These are normalized into weights, $\mathbf{w}_i = \mathrm{softmax}\left(\mathbf{s}_i\right)$. For each entity, the cumulative interactions are computed by the weighted mixture of all entities' value vectors, $\mathbf{a}_i = \sum_{j=1:N} w_{i,j} \mathbf{v}_j$. This can be compactly computed using matrix multiplications:
\begin{align}
    & A = \underbrace{\mathit{softmax}\left(\frac{Q K^T}{\sqrt{d}}\right)}_{\text{attention weights}} V
\end{align}

where $A$, $Q$, $K$, and $V$ compile the cumulative interactions, queries, keys, and values into matrices, and $d$ is the dimensionality of the key vectors used as a scaling factor. Like \cite{vaswani2017attention}, we use multiple, independent attention ``heads'', applied in parallel, which our attention visualisation analyses (see Results \ref{sec:box-world}) suggest may assume different relational semantics through training. The $\mathbf{a}^h_i$ vectors, where $h$ indexes the head, are concatenated together, passed to a multilayer perceptron ($2$-layer MLP with ReLU non-linearities) with the same layers sizes as $\mathbf{e}_i$, summed with $\mathbf{e}_i$ (i.e., a residual connection), and transformed via layer normalization \cite{ba2016layer}, to produce an output. Figure~\ref{fig:agent_bw} depicts this mechanism.

We refer to one application of this process as an ``attention block''. A single block performs non-local pairwise relational computations, analogous to relation networks \cite{santoro2017simple} and non-local neural networks \cite{wang2017non}. Multiple blocks with shared (recurrent) or unshared (deep) parameters can be composed to more easily approximate higher order relations, analogous to message-passing on graphs.

\subsubsection*{Extracting entities}

When dealing with unstructured inputs -- e.g., RGB pixels -- we need a mechanism to represent the relevant entities. We decide to make a minimal assumption that entities are things located in a particular point in space. We use a convolutional neural network (CNN) to parse pixel inputs into $k$ feature maps of size $n \times n$, where $k$ is the number of output channels of the CNN. We then concatenate $x$ and $y$ coordinates to each $k$-dimensional pixel feature-vector to indicate the pixel's position in the map. We treat the resulting $n^2$ pixel-feature vectors as the set of entities by compiling them into a $n^2 \times k$ matrix $E$. As in \cite{santoro2017simple}, this provides an efficient and flexible way to learn representations of the relevant entities, while being agnostic to what may constitute an entity for the particular problem at hand.

\subsubsection*{Agent architecture for Box-World}

We adopted an actor-critic set-up, using a distributed agent based on an Importance Weighted Actor-Learner Architecture \cite{espeholt2018impala}. The agent consists of $100$ actors, which generate trajectories of experience, and a single learner, which directly learns a policy $\pi$ and a baseline function $V$, using the actors' experiences. The model updates were performed on GPU using mini-batches of $32$ trajectories provided by the actors via a queue. 

The complete network architecture is as follows. The input observation is first processed through two convolutional layers with $12$ and $24$ kernels, $2 \times 2$ kernel sizes and a stride of $1$, followed by a rectified linear unit (ReLU) activation function. The output is tagged with two extra channels indicating the spatial position ($x$ and $y$) of each cell in the feature map using evenly spaced values between $-1$ and $1$. This is then passed to the \textit{relational module} (described above) consisting of a variable number of stacked MHDPA blocks, using shared weights. The output of the relational module is aggregated using feature-wise max-pooling across space (i.e., pooling a $n \times n \times k$ tensor to a $k$-dimensional vector), and finally passed to a small MLP to produce policy logits (normalized and used as multinomial distribution from which the action was sampled) and a baseline scalar $V$. 

Our baseline control agent replaces the MHDPA blocks with a variable number of residual convolution blocks. Please see the Appendix for further details, including hyperparameter choices.

\subsubsection*{Agent architecture for StarCraft II}

The same set-up was used for the StarCraft II agent, with a few differences in the network architecture to accommodate the specific requirements of the StarCraft II Learning Environment (SC2LE, \cite{vinyals2017starcraft}). In particular, we increased its capacity using 2 residual blocks, each consisting of 3 convolutional layers with $3 \times 3$ kernels, 32 channels and stride 1. We added a 2D-ConvLSTM immediately downstream of the residual blocks, to give the agent the ability to deal with recent history. We noticed that this was critical for StarCraft because the consequences of an agent's actions are not necessarily part of its future observations. For example, suppose the agent chooses to move a marine along a certain path at timestep $t$. At $t+\tau$ the agent's observation may depict the marine in a different location, 
but the details of the path are not depicted. In these situations, the agent is prone to re-select the path it had already chosen, rather than, say, move on to choose another action.

For the output, alongside action $a$ and value $V$, the network produces two sets of action-related arguments: non-spatial arguments ($\mathit{Args}$) and spatial arguments ($\mathit{Args}_{x,y}$). These arguments are used as modifiers of particular actions (see \cite{vinyals2017starcraft}). $\mathit{Args}$ are produced from the output of the aggregation function, whereas $\mathit{Args}_{x,y}$ result from upsampling the output of the relational module.

As in Box-World, our baseline control agent replaces the MHDPA blocks with a variable number of residual convolution blocks. Please see the Appendix for further details.

\section{Experiments and results}

\subsection{Box-World}\label{sec:box-world}

\subsubsection*{Task description}

\begin{figure}
    \centering
    \includegraphics[width=.9\textwidth]{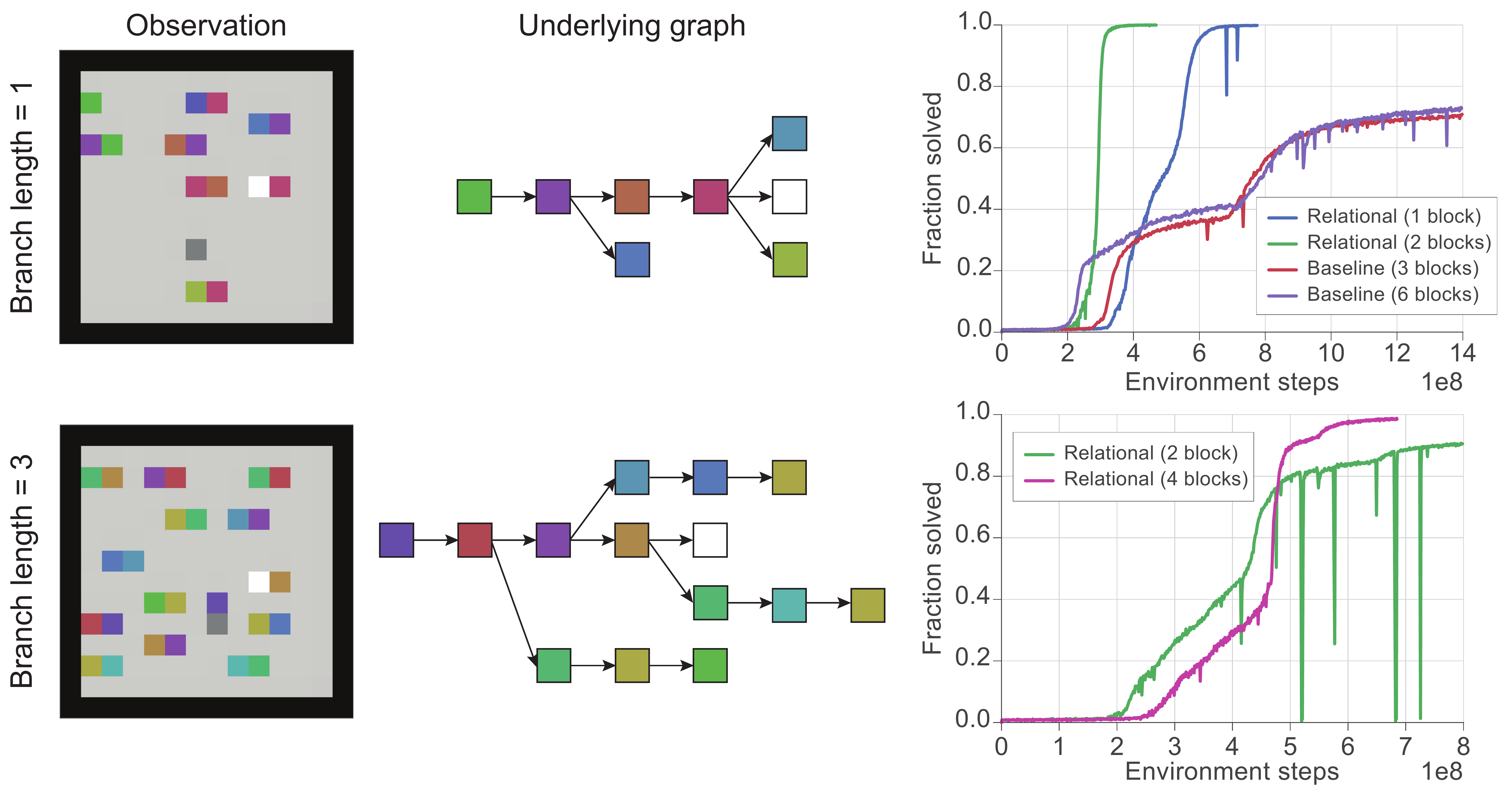}
    \caption{Box-World task: example observations (left), underlying graph structure that determines the proper path to the goal and any distractor branches (middle) and training curves (right).}
    \label{fig:boxworld}
\end{figure}

Box-World\footnote{The Box-World environment will be made publicly available online.} is a perceptually simple but combinatorially complex environment that requires abstract relational reasoning and planning. It consists of a $12 \times 12$ pixel room with keys and boxes randomly scattered. The room also contains an agent, represented by a single dark gray pixel, which can move in four directions: \textit{up}, \textit{down}, \textit{left}, \textit{right} (see Figure~\ref{fig:bw_sc_example}).

Keys are represented by a single colored pixel. The agent can pick up a loose key (i.e., one not adjacent to any other colored pixel) by walking over it. Boxes are represented by two adjacent colored pixels -- the pixel on the right represents the box's lock and its color indicates which key can be used to open that lock; the pixel on the left indicates the content of the box which is inaccessible while the box is locked.

To collect the content of a box the agent must first collect the key that opens the box (the one that matches the lock's color) and walk over the lock, which makes the lock disappear. At this point the content of the box becomes accessible and can be picked up by the agent. Most boxes contain keys that, if made accessible, can be used to open other boxes. One of the boxes contains a gem, represented by a single white pixel. The goal of the agent is to collect the gem by unlocking the box that contains it and picking it up by walking over it. Keys that an agent has in possession are depicted in the input observation as a pixel in the top-left corner.

In each level there is a unique sequence of boxes that need to be opened in order to reach the gem. Opening one wrong box (a distractor box) leads to a dead-end where the gem cannot be reached and the level becomes unsolvable.
There are three user-controlled parameters that contribute to the difficulty of the level: (1) the number of boxes in the path to the goal (solution length); (2) the number of distractor branches; (3) the length of the distractor branches. In general, the task is computationally difficult for a few reasons. First, a key can only be used once, so the agent must be able to reason about whether a particular box is along a distractor branch or along the solution path. Second, keys and boxes appear in random locations in the room, emphasising a capacity to reason about keys and boxes based on their abstract relations, rather than based on their spatial positions.

\subsubsection*{Training results}

The training set-up consisted of Box-World levels with solution lengths of at least 1 and up to 4. This ensured that an untrained agent would have a small probability of reaching the goal by chance, at least on some levels.\footnote{An agent with a random policy solves by chance 2.3\% of levels with solution lengths of 1 and 0.0\% of levels with solution lengths of 4.} The number of distractor branches was randomly sampled from 0 to 4. Training was split into two variants of the task: one with distractor branches of length 1; another one with distractor branches of length 3 (see Figure~\ref{fig:boxworld}). 

Agents augmented with our relational module achieved close to optimal performance in the two variants of this task, solving more than 98\% of the levels. In the task variant with short distractor branches an agent with a single attention block was able to achieve top performance.
In the variant with long distractor branches a greater number of attention blocks was required, consistent with the conjecture that more blocks allow higher-order relational computations.
In contrast, our control agents, which can only rely on convolutional and fully-connected layers, performed significantly worse, solving less than 75\% of the levels across the two task variants.

We repeated these experiments, this time with backward branching in the underlying graph used to generate the level. With backward branching the agent does not need to plan far into the future; when it is in possession of a key, a successful strategy is always to open the matching lock. In contrast, with forward branching the agent can use a key on the wrong lock (i.e. on a lock along a distractor branch). Thus, forward branching demands more complicated forward planning to determine the correct locks to open, in contrast to backward branching where an agent can adopt a more reactive policy, always opting to open the lock that matches the key in possession (see Figure~\ref{fig:bw_backwards} in Appendix).

\subsubsection*{Visualization of attention weights}

\begin{figure}
    \centering
    \includegraphics[width=.85\textwidth]{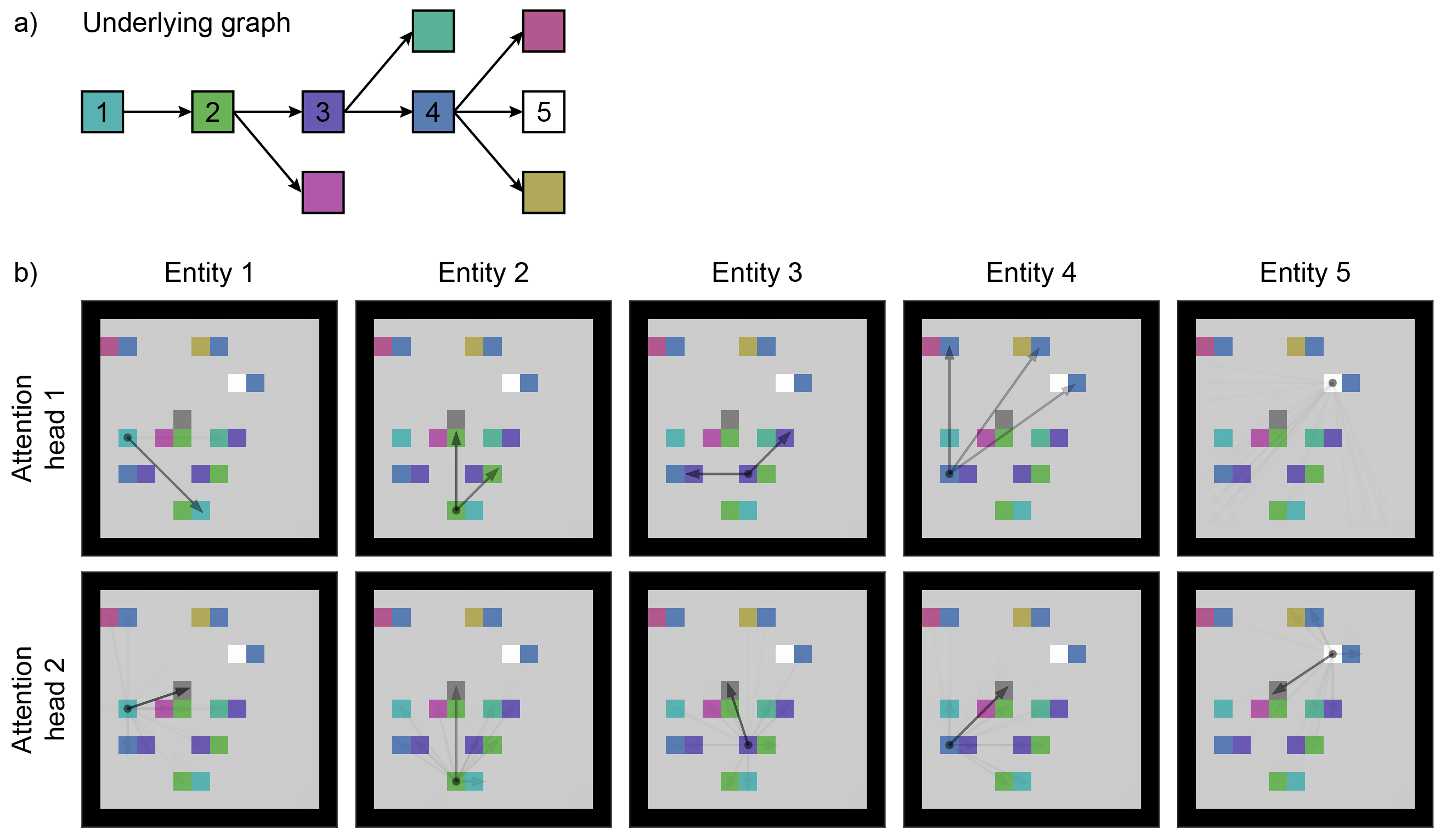}
    \caption{Visualization of attention weights. (a) The underlying graph of one example level; (b) the result of the analysis for that level, using each of the entities along the solution path (1--5) as the \textit{source} of attention. Arrows point to the entities that the \textit{source} is attending to. An arrow's transparency is determined by the corresponding attention weight.}
    \label{fig:attention}
\end{figure}

We next looked at specific rows of the matrix produced by $\mathit{softmax}(\frac{Q K^T}{\sqrt{d}})$; specifically, those rows mapping onto to relevant objects in the observation space. Figure \ref{fig:attention} shows the result of this analysis when the attending entities (source of the attention) are objects along the solution path. For one of the attention heads, each key attends mostly to the locks that can be unlocked with that key. In other words, the attention weights reflect the options available to the agent once a key is collected. For another attention head, each key attends mostly to the agent icon. This suggests that it is relevant to relate each object with the agent, which may, for example, provide a measure of relative position and thus influence the agent's navigation.

In the case of RGB pixel inputs, the relationship between keys and locks that can be opened with that key is confounded with the fact that keys and the corresponding locks have the same RGB representation. We therefore repeated the analysis, this time using one-hot representation of the input, where the mapping between keys and the corresponding locks is arbitrary. We found evidence for the following: (1) keys attend to the locks they can unlock; (2) locks attend to the keys that can be used to unlock them; (3) all the objects attend to the agent location; (4) agent and gem attend to each other and themselves.

\subsubsection*{Generalization capability: testing on withheld environments}

As we observed, the attention weights captured a link between a key and its corresponding lock, using a shared computation across entities. If the function used to compute the weights (and hence, used to determine that certain keys and locks are related) has learned to represent some general, abstract notion of what it means to ``unlock'' -- e.g., \texttt{unlocks(key, lock)} -- then this function should be able to generalize to key-lock combinations that it has never observed during training. Similarly, a capacity to understand ``unlocking'' shouldn't necessarily be affected by the number of locks that need to be unlocked to reach a solution. 

And so, we tested the model under two conditions, \textit{without further training}: (1) on levels that required opening a longer sequence of boxes than it had ever observed (6, 8 and 10), and (2) on levels that required using a key-lock combination that was never required for reaching the gem during training, instead only being placed on distractor paths. In the first condition the agent with the relational module solved more than 88\% of the levels, across all three solution length conditions. In contrast, the agent trained without the relational module had its performance collapse to 5\% when tested on sequences of 6 boxes and to 0\% on sequences of 8 and 10. On levels with new key-lock combinations, the agent augmented with a relational module solved 97\% of the new levels. The agent without the relational module performed poorly, reaching only 13\%. Together, these results show that the relational module confers on our agents, at least to a certain extent, the ability to do zero-shot transfer to more complex and previously unseen problems, a skill that so far has been difficult to attain using neural networks.

\begin{figure}
    \centering
    \includegraphics[width=.85\textwidth]{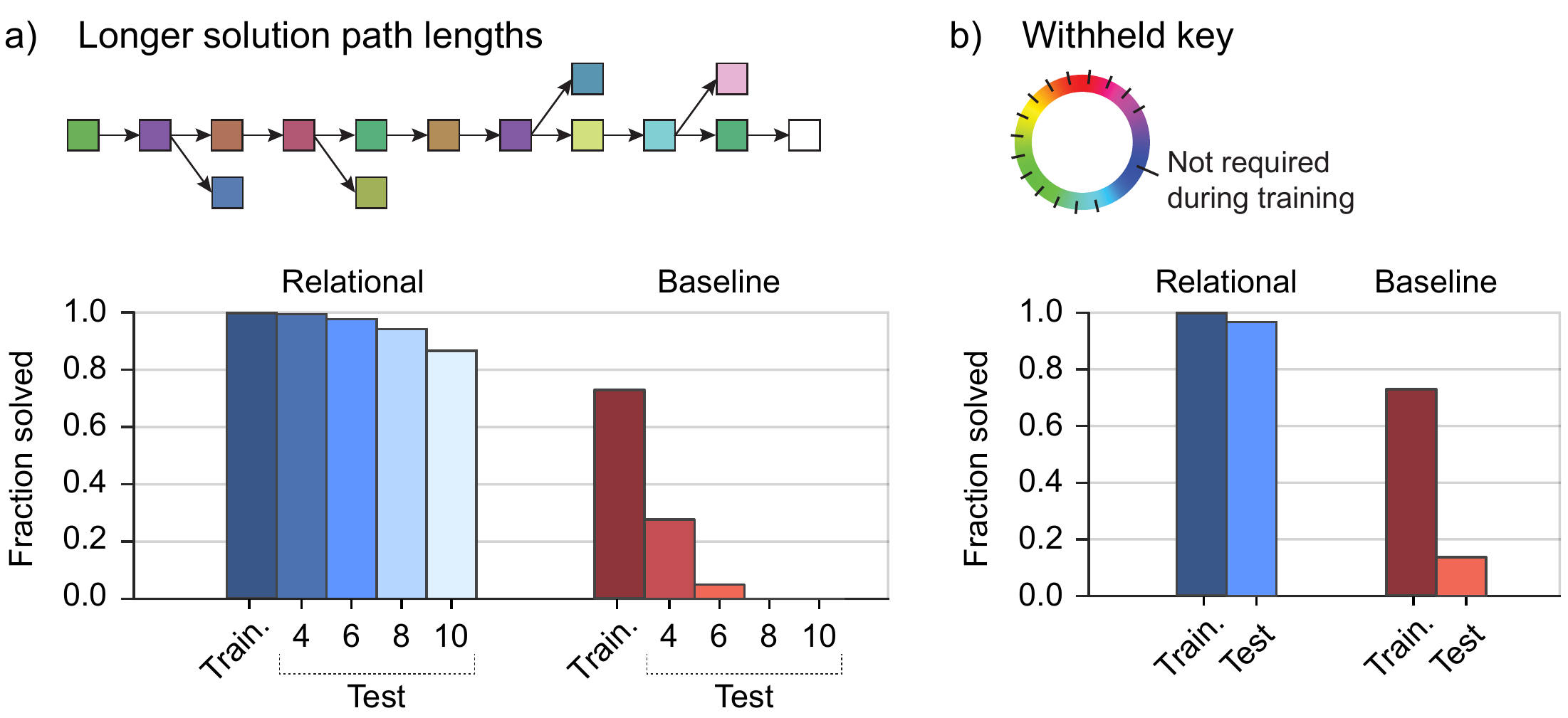}
    \caption{Generalization in Box-World. Zero-shot transfer to levels that required: (a) opening a longer sequence of boxes; (b) using a key-lock combination that was never required during training.}
    \label{fig:generalization}
\end{figure}

\subsection{StarCraft II mini-games}

\subsubsection*{Task description}

StarCraft II is a popular video game that presents a very hard challenge for reinforcement learning. It is a multi-agent game where each player controls a large number (hundreds) of units that need to interact and collaborate (see Figure \ref{fig:bw_sc_example}). It is partially observable and has a large action space, with more than 100 possible actions. The consequences of any single action -- in particular, early decisions in the game -- are typically only observed many frames later, posing difficulties in temporal credit assignment and exploration.

We trained our agents on the suite of 7 mini-games developed for the StarCraft II Learning Environment (SC2LE, \cite{vinyals2017starcraft}). These mini-games were proposed as a set of specific scenarios that are representative of the mechanics of the full game and can be used to test agents in a simpler set up with a better defined reward structure, compared to the full game.

\renewcommand{\arraystretch}{1.5}
\begin{table}
\centering
\small
\begin{threeparttable}
\begin{tabular}{ l c c c c c c c } & \multicolumn{7}{c}{\textbf{Mini-game}} \\
\textbf{Agent} & \circled{1} & \circled{2} & \circled{3} & \circled{4} & \circled{5} & \circled{6} & \circled{7} \\[0.1cm]
\hline\hline
DeepMind Human Player \cite{vinyals2017starcraft} 
        & 26 & 133 & 46 & 41 & 729 & 6880 & 138 \\
StarCraft Grandmaster \cite{vinyals2017starcraft}
        & 28 & 177 & 61 & 215 & 727 & 7566 & 133 \\
\hline
Random Policy \cite{vinyals2017starcraft}
        & 1 & 17 & 4 & 1 & 23 & 12 & < 1 \\
FullyConv LSTM \cite{vinyals2017starcraft}
        & 26 & 104 & 44 & 98 & 96 & 3351 & 6 \\
PBT-A3C \cite{jaderberg2017population}
        & -- & 101 & 50 & 132 & 125 & 3345 & 0 \\
\hline
Relational agent
        & \textbf{27} & \textbf{196} $\uparrow$ & \textbf{62} $\uparrow$ & \textbf{303} $\uparrow$ & \textbf{736} $\uparrow$ & 4906 & \textbf{123} \\
Control agent
        & \textbf{27} & 187 $\uparrow$ & 61 & 295 $\uparrow$ & 602 & \textbf{5055} & 120 \\
\hline\hline
\addlinespace[0.2cm]
\end{tabular}
\end{threeparttable}
\caption{Mean scores achieved in the StarCraft II mini-games using full action set. $\uparrow$ denotes a score that is higher than a StarCraft Grandmaster. Mini-games: (1) Move To Beacon, (2) Collect Mineral Shards, (3) Find And Defeat Zerglings, (4) Defeat Roaches, (5) Defeat Zerglings And Banelings, (6) Collect Minerals And Gas, (7) Build Marines.}
\label{table:main}
\end{table}

\subsubsection*{Training results}

For these results we used the full action set provided by SC2LE and performance was measured as the mean score over 30 episodes for each mini-game. Our agent implementations achieved high scores across all the mini-games (Table~\ref{table:main}). In particular, the agent augmented with a relational module achieved state-of-the-art results in six mini-games and its performance surpassed that of the human grandmaster in four of them.\footnote{For replay videos visit: \mbox{\url{http://bit.ly/2kQWMzE}}}

Head-to-head comparisons between our two implementations show that the agent with the relational component (relational) achieves equal or better results than the one without (control) across all mini-games. We note that both models improved substantially over the previous best \cite{vinyals2017starcraft}. This can be attributed to a number of factors: better RL algorithm \cite{espeholt2018impala}, better hyperparameter tuning to address issues of credit assignment and exploration, longer training, improved architecture, and a different action selection procedure. Next, we focus on differences afforded by relational inductive biases and turn to particular generalization tests to determine the behavioural traits of the control and relational agents.

\subsubsection*{Generalization capability}

As observed in Box-World, a capacity to better understand underlying relational structure -- rather than latch onto superficial statistics -- may manifest in better generalization to never-before-seen situations. To test generalization in SC2 we took agents trained on Collect Mineral Shards, which involved using two marines to collect randomly scattered minerals and tested them, without further training, on modified levels that allowed the agents to instead control \textit{five} marines. Intuitively, if an agent understands that marines are independent units that can be coordinated, yet controlled independently to collect resources, then increasing the number of marines available should only affect the underlying strategy of unit deployment, and should not catastrophically break model performance. 

We observed that -- at least for medium size networks -- there may be some interesting generalization capabilities, with the best seed of the relational agent achieving better generalization scores in the test scenario. However, we noticed high variability in these results, with the effect diminishing when using larger models (which may be more prone to overfitting on the training set). Therefore, more work is needed to understand the generalization effects of using a relational agent in StarCraft II (see Figure~\ref{fig:sc_generalization} in Appendix).

Given the combinatoric richness of the full-game, an agent is frequently exposed to situations on which it was not trained. Thus, an improved capacity to generalize to new situations caused by a better understanding of underlying, abstract relations is important.

\section{Conclusion}

By introducing structured perception and relational reasoning into deep RL architectures, our agents can learn interpretable representations, and exceed baseline agents in terms of sample complexity, ability to generalize, and overall performance. This demonstrates key benefits of marrying insights from RRL with the representational power of deep learning. Instead of trying to directly characterize the internal representations, we appealed to: (1) a behavioural analysis, and (2) an analysis of the internal mechanisms of the attention mechanism we used to compute entity-entity interactions. (1) showed that the learned representations allowed for better generalization, which is characteristic of relational representations. (2) showed that the model's internal computations were interpretable, and congruent with the computations we would expect from a model computing task-relevant relations.

Future work could draw on computer vision for more sophisticated structured perceptual reasoning mechanisms (e.g., \cite{chen2018iterative}), and hierarchical RL and planning \cite{vezhnevets2017feudal,guez2018learning} to allow structured representations and reasoning to translate more fully into structured behaviors. It will also be important to further explore the semantics of the agent's learned representations, through the lens of what one might hard-code in traditional RRL.

More speculatively, this work blurs the line between model-free agents, and those with a capacity for more abstract planning. An important feature of model-based approaches is making general knowledge of the environment available for decision-making. Here our inductive biases for entity- and relation-centric representations and iterated reasoning reflect key knowledge about the structure of the world. While not a model in the technical sense, it is possible that the agent learns to exploit this relational architectural prior similarly to how an imagination-based agent's forward model operates \cite{hamrick2017metacontrol,pascanu2017learning,weber2017imagination}.
More generally, our work opens new directions for RL via a principled hybrid of flexible statistical learning and more structured approaches.

\endgroup

\vskip 0.5in
\subsection*{Acknowledgments}
We would like to thank Richard Evans, Th\'{e}ophane Weber, Andr\'{e} Barreto, Daan Wierstra, John Agapiou, Petko Georgiev, Heinrich Küttler, Andrew Dudzik, Aja Huang, Ivo Danihelka, Timo Ewalds and many others on the DeepMind team.

\clearpage
\small
\bibliographystyle{unsrt}
\bibliography{bibliography2}

\renewcommand{\thesubsection}{\Alph{subsection}}
\appendix
\section*{Appendix}

\subsection{Box-world}

\subsubsection*{Task}
Each level in Box-world is procedurally generated. We start by generating a random graph (a tree) that defines the correct path to the goal -- i.e., the sequence of boxes that need to be opened to reach the gem. This graph also defines multiple distractor branches -- boxes that lead to dead-ends. The agent, keys and boxes, including the one containing the gem, are positioned randomly in the room, assuring that there is enough space for the agent to navigate between boxes. There is a total of 20 keys and 20 locks that are randomly sampled to produce the level. An agent receives a reward of $+10$ for collecting the gem, $+1$ for opening a box in the solution path and $-1$ for opening a distractor box. A level terminates immediately after the gem is collected or a distractor box is opened.

The generation process produces a very large number of possible trees, making it extremely unlikely that the agent will face the same level twice. The procedural generation of levels also allows us to create different training-test splits by withholding levels that conform to a particular case during training and presenting them to the agent at test time.

\subsubsection*{Agent architecture}

The agent had an entropy cost of $0.005$, discount ($\gamma$) of $0.99$ and unroll length of $40$ steps. Queries, keys and values were produced by 2 to 4 attention heads and had an embedding size ($d$) of 64. The output of this module was aggregated using a feature-wise max pooling function and passed to a 4 fully connected layers, each followed by a ReLU. Policy logits ($\pi$, size 4) and baseline function ($V$, size 1) were produced by a linear projection. The policy logits were normalized and used as multinomial distribution from which the action ($a$) was sampled.

Training was done using RMSprop optimiser with momentum of 0, $\epsilon$ of $0.1$ and a decay term of $0.99$. The learning rate was tuned, taking values between $1\mathrm{e}{-5}$ and $2\mathrm{e}{-4}$. Informally, we note that we could replicate these results using an A3C setup, though training took longer.

\subsubsection*{Control agent architecture}

As a baseline control agent we used the same architecture as the relational agent but replaced the relational module with a variable number (3 to 6) of residual-convolutional blocks. Each residual block comprised two convolutional layers, with $3 \times 3$ kernels, stride of 1 and 26 output channels.

\begin{figure}[b]
    \centering
    \includegraphics[width=.8\textwidth]{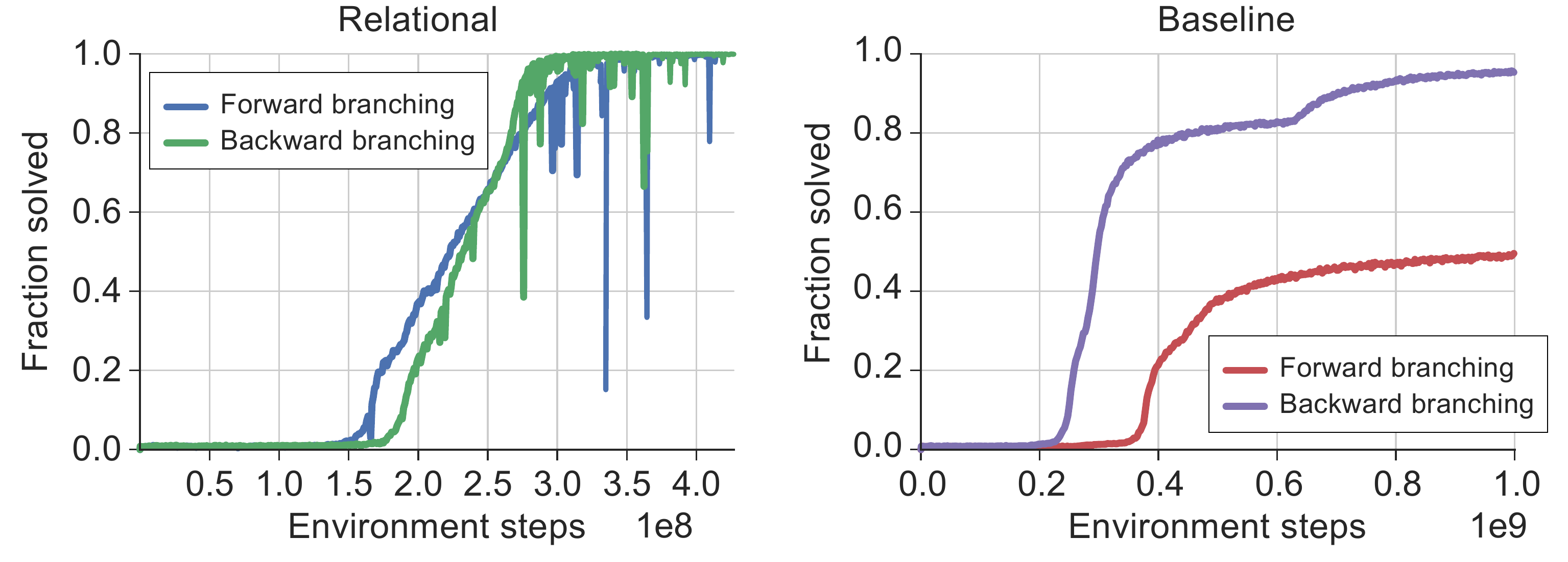}
    \caption{Box-World: forward branching versus backward branching. With backward branching, any given key can only open one box; however, each key type (i.e. color), can appear in multiple boxes. This means that an agent can adopt a more reactive policy without planning beyond which box to open next.}
    \label{fig:bw_backwards}
\end{figure}

\subsection{StarCraft II mini-games}

Starcraft II agents were trained with Adam optimiser for a total of 10 billion steps using batches of 32 trajectories, each unrolled for 80 steps. A linear decay was applied to the optimiser learning rate and entropy loss scaling throughout training (see Table~\ref{table:sc2_fixed_hp} for details). We ran approximately 100 experiments for each mini-game, following Table~\ref{table:sc2_rel_hp} hyperparameter settings and 3 seeds.

\begin{figure}
    \centering
    \includegraphics[width=.7\textwidth]{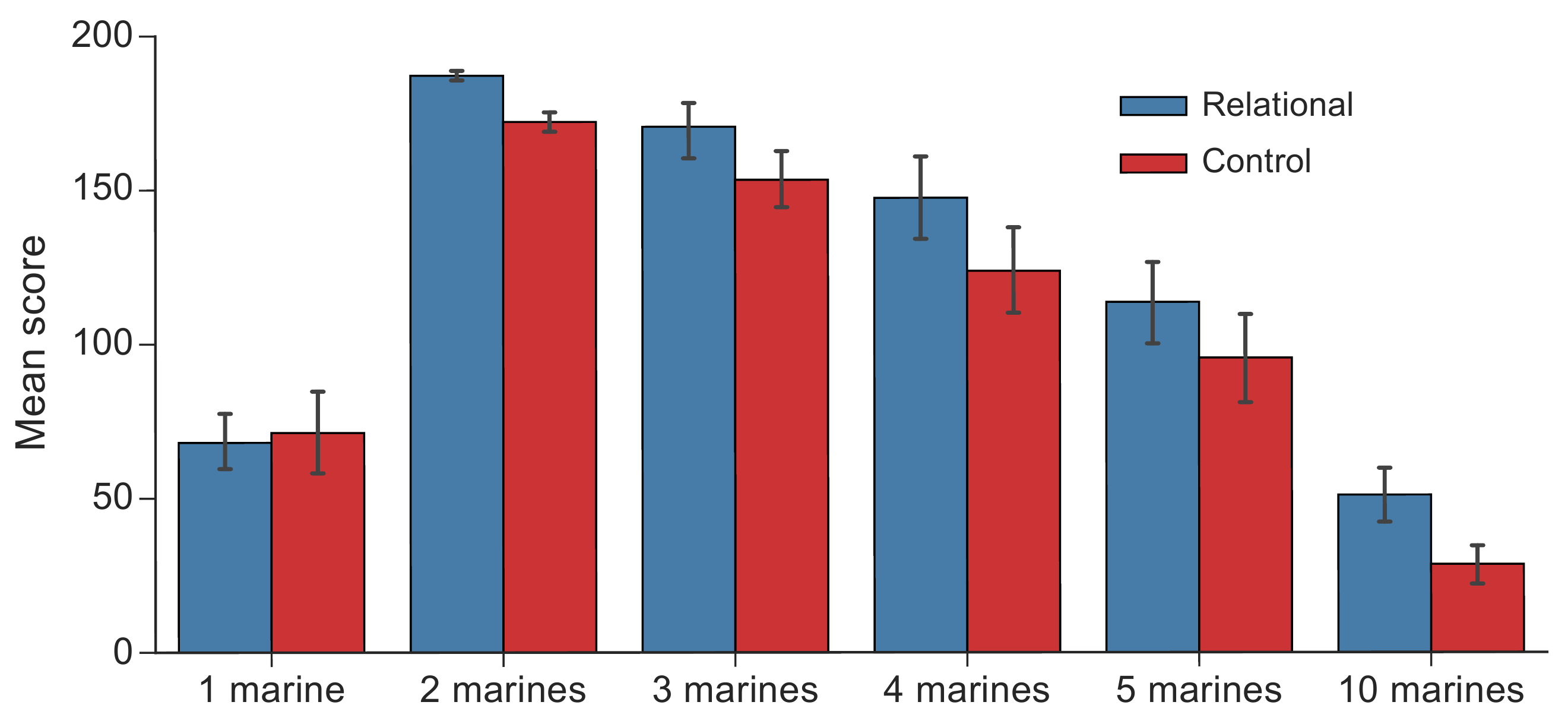}
    \caption{Generalization results on the StarCraft II mini-game Collect Mineral Shards. Agents were trained on levels with 2 marines and tested on levels with 1, 2, 3, 4, 5 or 10 marines. Colored bars indicate mean score of the ten best seeds; error bars indicate standard error.}
    \label{fig:sc_generalization}
\end{figure}

\subsubsection*{Relational Agent architecture}

The StarCraft II (SC2) agent architecture follows closely the one we adopted in Box-World. Here we highlight the changes needed to satisfy SC2 constraints.

\textbf{Input-preprocessing}. At each time step agents are presented with 4 sources of information: \textit{minimap}, \textit{screen}, \textit{player}, and \textit{last-action}. These tensors share the same pre-processing: numerical features are rescaled with a logarithmic transformation and categorical features are embedded into a continuous 10-dimensional space.

\textbf{State encoding}. Spatially encoded inputs (\textit{minimap} and \textit{screen}) are tiled with binary masks denoting whether the previous action constituted a screen- or minimap-related action. These tensors are then fed to independent residual convolutional blocks, each consisting of one convolutional layer ($4 \times 4$ kernels and stride 2) followed by a residual block with 2 convolutional layers ($3 \times 3$ kernels and stride 1), which process and downsample the inputs to [$8 \times 8 \times \#\mathit{channels}_{1}$] outputs. These tensors are concatenated along the depth dimension to form a singular spatial input ($\mathit{inputs}_{3D}$). The remaining inputs (\textit{player} and \textit{last-action}) are concatenated and passed to a 2-layer MLP (128 units, ReLU, 64 units) to form a singular non-spatial input ($\mathit{inputs}_{2D}$).

\textbf{Memory processing}. Next, $\mathit{inputs}_{2D}$ is passed to the Conv2DLSTM along with its previous state to produce a new state and $\mathit{outputs}_{2D}$, which represents an aggregated history of input observations.

\textbf{Relational processing}. $\mathit{outputs}_{2D}$ is flattened and passed to the stacked MHDPA blocks (see Table~\ref{table:sc2_relational_settings} for details). Its output tensors follow two separate pathways -- \textit{relational-spatial}: reshapes the tensors to their original spatial shape [$8 \times 8 \times \#\mathit{channels}_{2}$]; \textit{relational-nonspatial}: aggregates through a feature-wise max-pooling operation and further processes using a 2-layer MLP (512 units per layer, ReLU activations). 

\textbf{Output processing}. $\mathit{inputs}_{2D}$ and \textit{relational-nonspatial} are concatenated to form a set of \textit{shared features}. Policy logits are produced by feeding \textit{shared features} to a 2-layer MLP (256 units, ReLU, $|actions|$ units) and masking unavailable actions (following \cite{vinyals2017starcraft}). Similarly, baselines values $V$ are generated by feeding \emph{shared features} to a separate 2-layer MLP (256 units, ReLU, 1 unit). 

Actions are sampled using computed policy logits and embedded into a 16 dimensional vector. This embedding is used to condition \emph{shared features} and generate logits for non-spatial arguments ($\mathit{Args}$) through independent linear combinations (one for each argument). Finally, spatial arguments ($\mathit{Args}_{x,y}$) are obtained by first deconvolving \textit{relational-spatial} to [$32 \times 32 \times \#\mathit{channels}_{3}$] tensors using Conv2DTranspose layers, conditioned by tiling the action embedding along the depth dimension and passed to a $1 \times 1 \times 1$ convolution layers (one for each spatial argument). Spatial arguments ($x$, $y$) are produced by sampling resulting tensors and selecting the corresponding row and column indexes.

\subsubsection*{Control agent architecture}

The baseline control agent architecture only differs on the \emph{relational processing} part of the pipeline. Analogous to the relational agent, $outputs_{2D}$ are obtained from Conv2DLSTM layers. These tensors are first passed to a 12-layer deep residual model -- comprising 4 blocks of 3 convolutions layers (32 output channels, $4 \times 4$ kernel for the first convolution and $3 \times 3$ for the second and third, and stride 1) interleaved with ReLU activations and skip-connections -- as proposed by \cite{he2016identity}, to form the \emph{relational-spatial} outputs. These tensors also follow a separate pathway where they are flattened and passed to a 2-layer MLP (512 units per layer, ReLU activations) to produce what we refer to above as \emph{relational-nonspatial}. The remaining architecture is identical to the relational agent.

\begin{table}
\centering
\small
\begin{tabular}{ l c r }
\hline
\textbf{Hyperparameter} & & \textbf{Value} \\
\hline\hline
Conv2DLSTM      & &  \\
\hspace{5mm} Output channels ($\#\mathit{channels}_{1}$) & & 96 \\
\hspace{5mm} Kernel shape    & & (3, 3) \\
\hspace{5mm} Stride          & & (1, 1) \\
Conv2DTranspose & & \\
\hspace{5mm} Output channels ($\#\mathit{channels}_{3}$) & & 16 \\ 
\hspace{5mm} Kernel shape    & & (4, 4) \\
\hspace{5mm} Stride          & & (2, 2) \\
Discount ($\gamma$)          & & 0.99 \\
Batch size                   & & 32 \\
Unroll Length                & & 80 \\
Baseline loss scaling        & & 0.1 \\
Clip global gradient norm    & & 100.0 \\
Adam $\beta_1$               & & 0.9 \\
Adam $\beta_2$               & & 0.999 \\
Adam $\epsilon$              & & $1\mathrm{e}{-8}$ \\
\hline
\addlinespace[0.2cm]
\end{tabular}
\caption{Shared fixed hyperparameters across mini-games.}
\label{table:sc2_fixed_hp}
\end{table}

\begin{table}
\centering
\small
\begin{tabular}{ l c r }
\hline
\textbf{Setting} & & \textbf{Value} \\
\hline\hline
MLP layers   & & 2 \\
Units per MLP layer   & & 384 \\
MLP activations   & & ReLU \\
Attention embedding size & & 32 \\
Weight sharing  & & shared MLP across blocks \\
                & & shared embedding across blocks \\
\hline
\addlinespace[0.2cm]
\end{tabular}
\caption{Fixed MHDPA settings for StarCraft II mini-games.}
\label{table:sc2_relational_settings}
\end{table}

\begin{table}
\centering
\small
\begin{tabular}{ l c r }
\hline
\textbf{Hyperparameter} & & \textbf{Value} \\
\hline\hline
Relational module  & &  \\
\hspace{5mm} Number of heads   & & [1, 3]\\
\hspace{5mm} Number of blocks   & & [1, 3, 5]\\
Entropy loss scaling         & & [$1\mathrm{e}{-1}, 1\mathrm{e}{-2}, 1\mathrm{e}{-3}]$ \\
Adam learning rate           & & [$1\mathrm{e}{-4}, 1\mathrm{e}{-5}]$ \\
\hline
\addlinespace[0.2cm]
\end{tabular}
\caption{Swept hyperparameters across mini-games.}
\label{table:sc2_rel_hp}
\end{table}

\end{document}